\documentclass[conference]{IEEEtran}
\IEEEoverridecommandlockouts

\usepackage{cite}
\usepackage{amsmath,amssymb,amsfonts}
\usepackage{graphicx}
\usepackage{textcomp}
\usepackage{xcolor}
\def\BibTeX{{\rm B\kern-.05em{\sc i\kern-.025em b}\kern-.08em
    T\kern-.1667em\lower.7ex\hbox{E}\kern-.125emX}}

\usepackage{comment} %
\usepackage{etoolbox} %
\usepackage{pdflscape} %
\usepackage[linesnumbered,ruled,vlined]{algorithm2e} %

\SetCommentSty{commentfont}
\usepackage{url} %
\usepackage[pagebackref,breaklinks]{hyperref} %
\usepackage[capitalize]{cleveref} %

\usepackage[caption=false]{subfig} %

\usepackage{paralist} %

\usepackage{booktabs} %
\usepackage{multirow} %
\usepackage{multicol} %
\usepackage{longtable} %
\newcommand{\itseries}{\fontshape{it}\selectfont} %
\robustify\itseries
\newrobustcmd{\IT}{\itseries} 

\usepackage{pifont} %

\newcommand{\x}{\mathbf{x}}
\renewcommand{\c}{\mathbf{c}}

\newcommand{\cls}{\texttt{CLS} cos\_sim}

\usepackage{xspace}
\newcommand{\perf}{performance\xspace}

\begin{document}
\bstctlcite{MyBSTcontrol} %

\title{
Beyond Classification: Evaluating Diffusion Denoised Smoothing for Security-Utility Trade off
}

\author{\IEEEauthorblockN{Yury Belousov, Brian Pulfer, Vitaliy Kinakh and Slava Voloshynovskiy}
	\IEEEauthorblockA{\textit{Department of Computer Science, University of Geneva, Switzerland} \\ \{yury.belousov, brian.pulfer, vitaliy.kinakh, svolos\}@unige.ch}
}

\maketitle

\begin{abstract}
    While foundation models demonstrate impressive performance across various tasks, they remain vulnerable to adversarial inputs. Current research explores various approaches to enhance model robustness, with Diffusion Denoised Smoothing emerging as a particularly promising technique. This method employs a pretrained diffusion model to preprocess inputs before model inference. Yet, its effectiveness remains largely unexplored beyond classification. We aim to address this gap by analyzing three datasets with four distinct downstream tasks under three different adversarial attack algorithms. Our findings reveal that while foundation models maintain resilience against conventional transformations, applying high-noise diffusion denoising to clean images without any distortions significantly degrades performance by as high as 57\%. Low-noise diffusion settings preserve performance but fail to provide adequate protection across all attack types. Moreover, we introduce a novel attack strategy specifically targeting the diffusion process itself, capable of circumventing defenses in the low-noise regime. Our results suggest that the trade-off between adversarial robustness and performance remains a challenge to be addressed. 

\end{abstract}


\section{Introduction}

Vision Foundation Models (VFMs) have transformed computer vision across diverse tasks. Despite their impressive capabilities, these models remain highly vulnerable to adversarial perturbations, i.e., imperceptible modifications that can catastrophically degrade the performance of downstream tasks.

Recent approaches like "(Certified!!) Adversarial Robustness for Free!"~\cite{carlini2023free} propose using pre-trained diffusion models to denoise inputs before inference. However, existing research primarily focuses on classification tasks with traditional neural networks rather than examining foundation models across multiple applications.

We present the first comprehensive evaluation of diffusion-based defenses for VFMs across multiple downstream tasks. Our analysis reveals: 
\begin{inparaenum}[(i)]
    \item A stark contrast between VFMs' resilience to common image distortions versus their extreme vulnerability to adversarial attacks. 
    \item A critical trade-off: high-noise diffusion offers substantial protection but degrades clean performance by 14-33\% for classification, segmentation and retrieval tasks, and up to 57\% for depth estimation, while low-noise diffusion preserves performance but remains vulnerable to sophisticated attacks.
\end{inparaenum}

These findings highlight the persistent challenge of balancing strong adversarial robustness with high task performance in security-critical applications of foundation models.

{
\newcommand{\IMGSIZE}{0.1}

\newcommand{\imagesrow}[3]{
    \subfloat[clean]{\includegraphics[height=\IMGSIZE\textwidth]{Images/examples/clean_#1}}
    \quad 
    \subfloat[classified:\\ \texttt{#2}]{\includegraphics[height=\IMGSIZE\textwidth]{Images/examples/clean_seg_#1}}
    \hspace{2.2em}
    \subfloat[denoised]{\includegraphics[height=\IMGSIZE\textwidth]{Images/examples/noise_#1}}
    \quad 
    \subfloat[classified:\\  \texttt{#3}]{\includegraphics[height=\IMGSIZE\textwidth]{Images/examples/noise_seg_#1}}
    \qquad
}    

\captionsetup[subfigure]{labelformat=empty, justification=centering}

\begin{figure}[t!]
    \centering

    \imagesrow{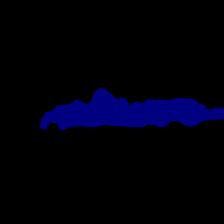}{boat}{aeroplane} \newline
    \imagesrow{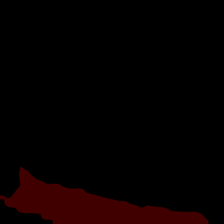}{cat}{train}

    \caption{Clean images before and after diffusion denoising with their corresponding predictions. Note that denoised images are distorted, and their classification labels and segmentation maps are incorrect.}
    \label{fig:denoising_examples}
    
\end{figure}
}

\section{Background}
\subsection{Vision Foundation Models}
VFMs are large-scale pre-trained neural networks capable of capturing rich, transferable visual features that can be adapted to numerous downstream tasks. However, this property introduces potential vulnerabilities, as adversarial perturbations targeting the VFM can propagate to multiple downstream tasks simultaneously, amplifying their harmful impact. %

\subsection{Adversarial Attacks and Robustness}

Adversarial attacks are imperceptible perturbations to input data that cause machine learning models to make incorrect predictions \cite{goodfellow2014explaining, szegedy2013intriguing}. Common attack methods include Projected Gradient Descent (PGD) \cite{madry18pgd}; Momentum Iterative Fast Gradient Sign Method (MI-FGSM) \cite{dong2018mifgsm}; and Scale-Invariant Attack (SIA) \cite{wang2023sia}.

Defense strategies include adversarial training \cite{madry18pgd} and input transformations \cite{guo2017countering}, with each approach balancing robustness against clean accuracy. 

\subsection{Randomized Smoothing and Denoised Smoothing}
Randomized smoothing is a technique that transforms a base classifier into a more robust version by taking a majority vote over multiple Gaussian-noised copies of an input. \cite{cohen2019certified} demonstrated that by adding Gaussian noise to an input and averaging classifier predictions, one can derive a certified robustness radius under the $\ell_2$-norm.

Denoised smoothing extends this concept by introducing an explicit denoising step before classification. \cite{salman2020denoised} proposed a two-step approach where an input is first corrupted with Gaussian noise, then denoised using a learned model before being classified. This enables certified robustness without retraining the classifier itself.

\subsection{Certified Adversarial Robustness Using Diffusion Models}
The Diffusion Denoising Smoothing \cite{carlini2023free} leverages off-the-shelf denoising diffusion probabilistic models \cite{ho2020denoising} as denoisers within a certified smoothing pipeline. 
The methodology involves:
\begin{inparaenum}[(1)]
     \item Applying Gaussian noise $\boldsymbol{\delta}_t$ with the variance \(\sigma_t^2=\frac{1-\alpha_t}{\alpha_t}\) to an input ${\bf x}_o$, where $\alpha_t$ is a constant derived from the timestamp $t$ that determines the amount of noise to be added to the image, resulting in \({\bf x}_{t}=\sqrt{\alpha_{t}}({\bf x}_o+\boldsymbol{\delta}_t), \quad \boldsymbol{\delta}_t \sim \mathcal{N}\left({\bf 0}, \sigma_t^2 \mathbf{I}\right)\). 
     \item Using a pretrained diffusion model $d_{\gamma}$ to denoise the perturbed input $\hat{{\bf x}}=d_{\gamma}\left( {\bf x}_{t} ; t \right)$.
     \item Classifying the denoised image using the output of the off-the-shelf VFM ${\bf z}=f_{\phi}(\hat{\bf x})$ and a downstream head ${\bf c} = g_{\theta}({\bf z})$, where $\bf c$ represents a downstream task label. 
\end{inparaenum}
     
In the original work \cite{carlini2023free}, the authors only considered the classification task.
Our work extends this evaluation by analyzing the impact of different noise levels, adversarial attacks, and downstream tasks, providing a broader perspective on certified protection mechanisms.

\section{Methodology}
\subsection{Adversarial Attacks}
Let us consider a model composed of a VFM  $f_{\phi}$ with a downstream task head $g_{\theta}$. For a given input $\x_o$, the system produces the output $\c_o = g_{\theta}(f_{\phi}(\x_o))$.

To attack the above system, the attacker defines two losses: $\mathcal{L}_x\left(\x; \x_o\right)$ measuring the perceptual distance between an input $\x$ and $\x_o$, and $\mathcal{L}_{c}(\c;\c_o)$
gauging how the task performance on $\x$ differs from the result $\c_o$.
For instance, for the classification task, the result $\c_o$ is a predicted class and $\mathcal{L}_{c}(\c;\c_o) = p(\c_o|\x) - \max_{\c\neq \c_o} p(\c|\x)$. %
Under a distortion budget constraint, the attack looks for the minimizer $\x_a$ of $\mathcal{L}_{c}(\c;\c_o)$ over $\{\x:\mathcal{L}_x\left(\x; \x_o\right)\leq\epsilon\}$, with $\c = g_{\theta}(f_{\phi}(\x))$.

\subsection{Adversarial attacks under denoising diffusion}
We outline a proposed general algorithm in \cref{algo:attack}. PGD~\cite{madry18pgd}, MI-FGSM~\cite{dong2018mifgsm}, and SIA~\cite{wang2023sia} can all be represented as a particular case. 
Since PGD does not use momentum, it is equivalent to setting decay factor $\mu$ to 0 for gradient accumulation. For MI-FGSM and SIA, decay factor $\mu$ is 1. PGD and MI-FGSM do not apply any transformation to the input, so the transformation function $\mathcal{T}$ is the identity function. For SIA,  $\mathcal{T}$ is a random transform out of resize, vertical or horizontal shift or flip, rotation by 180 degrees, scaling, adding Gaussian noise, or dropout with probability $0.1$.

When $p_{\text{diffusion}} = 0$, no denoising occurs and the original attack is used. When $p_{\text{diffusion}} = 1$, the attack uses the denoised image at every step via a single-step diffusion model using random timestep between $t_{\min}$ and $t_{\max}$.

\begin{algorithm}[t!]
\caption{Adversarial Attack with Certified Augmentations \label{algo:attack}}
\SetAlgoLined
\KwIn{
  $\mathbf{x}_o$: input image, $\mathbf{c}_o$: label; $f_{\phi}$: foundation model, $g_{\theta}$: downstream head; \\
  $\epsilon_{\infty}$: budget, $\nu$: step size, $T$: iterations, $\mu$: decay, $\mathcal{T}$: transform; \\
  $d_{\gamma}$: diffusion denoiser, $t_{\min}, t_{\max}$: denoising range, $p_{\text{diffusion}}$: denoise probability
}
\KwOut{$\mathbf{x}_a$: adversarial image}

$\mathbf{a} \leftarrow \mathbf{0}$ \tcp*[r]{Init perturbation} 
$\mathbf{g} \leftarrow \mathbf{0}$ \tcp*[r]{Init momentum}

\For{$t = 1$ \KwTo $T$}{
    $\mathbf{x}_a \leftarrow \mathbf{x}_o + \mathbf{a}$ \tcp*[r]{Perturbed image}
    \eIf{$\text{rand}() < p_{\text{diffusion}}$}{        
        $t \leftarrow \text{randint}(t_{\min}, t_{\max})$ \tcp*[r]{Sample timestep}
        ${\bf x}_{t} \leftarrow \sqrt{\alpha_{t}}({\bf x}_a+\boldsymbol{\delta}_t)$ \tcp*[r]{Noisy sample}
        $\hat{\mathbf{x}} \leftarrow d_{\gamma}({\bf x}_{t}, t)$ \tcp*[r]{Denoise}
    }{
        $\hat{\mathbf{x}} \leftarrow \mathbf{x}_a$ \tcp*[r]{No denoising}
    }
    $\mathbf{c} \leftarrow g_{\theta}(f_{\phi}(\mathcal{T}(\hat{\mathbf{x}})))$ \tcp*[r]{Predict}
    $\mathcal{L} \leftarrow \mathcal{L}_{c}(\mathbf{c};\mathbf{c}_o)$ \tcp*[r]{Loss}
    $\nabla_{\mathbf{a}}\mathcal{L} \leftarrow \text{Gradient}(\mathcal{L}, \mathbf{a})$ \tcp*[r]{Grad}
    $\mathbf{g} \leftarrow \mu \cdot \mathbf{g} + \frac{\nabla_{\mathbf{a}}\mathcal{L}}{\|\nabla_{\mathbf{a}}\mathcal{L}\|_1}$ \tcp*[r]{Update momentum}
    $\mathbf{a} \leftarrow \mathbf{a} + \nu \cdot \text{sign}(\mathbf{g})$ \tcp*[r]{Update perturbation}
    $\mathbf{a} \leftarrow \text{Clip}(\mathbf{a}, -\epsilon_{\infty}, \epsilon_{\infty})$ \tcp*[r]{Budget clip}
    $\mathbf{a} \leftarrow \text{Clip}(\mathbf{a}, -\mathbf{x}_o, 1-\mathbf{x}_o)$ \tcp*[r]{Valid range}
}
$\mathbf{x}_a \leftarrow \mathbf{x}_o + \mathbf{a}$

\Return{$\mathbf{x}_a$}
\end{algorithm}

\renewcommand{\paragraph}[1]{\vspace{1ex}\noindent\textbf{#1}}
\section{Results and Discussion}

\subsection{Experimental Setup}

\paragraph{Foundation Model.} For our study, we use the DINOv2~\cite{oquab2023dinov2} foundation model in three sizes: ViT-S, ViT-B and ViT-L \cite{dosovitskiy2021image}. We train a downstream head $g_{\theta}$ for each task using a train split and report performance on a validation or test part.

\paragraph{Diffusion denoising.} We employ an off-the-shelf unconditional denoising diffusion model~\cite{dhariwal2021diffusion}, which was trained on the ImageNet dataset~\cite{deng2009imagenet}. To address resolution compatibility requirements, we resize the smallest dimension of each image to 224 and then perform a central crop of $224 \times 224$. %

\paragraph{Datasets and Metrics. } 
\begin{inparaenum}[(i)]
    \item PascalVOC~\cite{pascal-voc-2012}: \textit{Accuracy} for classification and \textit{Mean Intersection over Union (mIoU)} for segmentation. 
    \item NYU-Depth~\cite{couprie2013indoor}: \textit{RMSE} for depth estimation.
    \item Revisited Oxford buildings dataset~\cite{RITAC18}: \textit{Mean Average Precision (mAP)} for image retrieval of easy queries.
\end{inparaenum} 
Additionally, we quantify the feature-space impact of adversarial perturbations by measuring the cosine similarity between the class tokens extracted from the last layer of the foundation model for original and adversarial images $\text{cos\_sim}(f_{\phi}(\x_o), f_{\phi}(\x_a))$ and referred to as \cls.

\paragraph{Adversarial Attacks.} We apply three attack methods -- PGD~\cite{madry18pgd}, MI-FGSM~\cite{dong2018mifgsm}, and SIA~\cite{wang2023sia} -- using consistent hyperparameters in \cref{algo:attack}.
The perturbation budget $\epsilon_{\infty}$ is set to $\frac{3}{255}$, with a Peak Signal-to-Noise Ratio \textit{PSNR} around 40 dB. We set the number of iterations $T$ to 50 and the step size $\nu$ to $\frac{\epsilon_{\infty}}{T} * 4$. For every downstream task we set $\mathcal{L}_{c}(\c;\c_o)$ to corresponding loss: cross-entropy for classification and segmentation, RMSE for depth estimation, and \cls{} for image retrieval. The denoising probability $p_{\text{diffusion}}$ during attacks is set to either 0, 0.5, or 1. When it is equal to zero it corresponds to the original attack method without using diffusion model at all. For the diffusion timestep parameter, we explore three distinct scenarios: (i) $t_{\min} = t_{\max} = 10$~-- corresponding to a low noise level; (ii) $t_{\min} = t_{\max} = 396$~-- corresponding to a high noise level; (iii) $t_{\min} = 10, \; t_{\max} = 396$~-- corresponding to a varying noise level.

To facilitate future research, the codebase will be made available upon paper acceptance.

\subsection{Robustness to Distortions and Impact of Model Size}
\begin{table*}[htbp]
  \caption{Downstream performance for DINOv2 under various distortions (\textit{robustness}), diffusion denoising, and adversarial attack (\textit{security}).}
  \label{tab:robustness_denoising}
  \centering
  \begin{tabular}{cccccccccc}
  \toprule
  \multirow{2}{*}{\textbf{Backbone}} & \multirow{2}{*}{\textbf{Transform}}& \multicolumn{3}{c}{\textbf{PascalVOC}} & \multicolumn{2}{c}{\textbf{NYU Depth}} & \multicolumn{2}{c}{\textbf{rOxford}} \\
  \cmidrule(lr){3-5} \cmidrule(lr){6-7} \cmidrule(lr){8-9}
   & & \textbf{Classification$\uparrow$} & \textbf{Segmentation$\uparrow$} & \textbf{\texttt{CLS} cos\_sim$\uparrow$} & \textbf{Depth$\downarrow$} & \textbf{\texttt{CLS} cos\_sim$\uparrow$} & \textbf{mAP$\uparrow$} & \textbf{\texttt{CLS} cos\_sim$\uparrow$} \\
  
  \midrule
  \multirow{10}{*}{ViT-S} & \IT clean performance & \IT 94.89 & \IT 79.26 & \IT 1.00 & \IT 0.55 & \IT 1.00 & \IT 84.46 & \IT 1.00 \\
  \cmidrule(lr){2-9}
    & hflip & 94.76 & 79.07 & 0.93 & 0.56 & 0.90 & 85.49 & 0.97 \\
    & wiener $\text{size}=13$ & 92.92 & 71.68 & 0.85 & 0.67 & 0.76 & 82.24 & 0.84 \\
    & blur $\text{kernel\_size}=13$ & 94.50 & 76.14 & 0.88 & 0.65 & 0.91 & 84.91 & 0.88 \\
    & jpeg $\text{quality}=85$ & 94.63 & 78.56 & 0.98 & 0.57 & 0.98 & 84.62 & 0.98 \\
    & grayscale & 94.36 & 78.71 & 0.94 & 0.59 & 0.96 & 83.84 & 0.96 \\
    & rotation & 82.57 & 59.49 & 0.57 & 1.16 & 0.44 & 28.55 & 0.40 \\
    & low noise diffusion & 95.54 & 79.03 & 0.99 & 0.58 & 0.99 & 83.82 & 0.99 \\
    & high noise diffusion & 80.60 & 61.30 & 0.60 & 0.79 & 0.58 & 62.03 & 0.61 \\
    & PGD Adversarial Attack & 0.0 & 10.5 & 0.2 & 6.41 & 0.14 & 0.46 & -0.62 \\
  
  \midrule
  \multirow{3}{*}{ViT-B} & \IT clean performance & \IT 96.99 & \IT 81.02 & \IT 1.00 & \IT 0.47  & \IT 1.00 & \IT 88.17  & \IT 1.00 \\     
  & low noise diffusion & 96.59 & 81.15 & 0.99 & 0.47  & 0.99  & 88.13  & 0.99 \\
  & high noise diffusion & 82.44 & 66.35 & 0.55 & 0.71  & 0.53  & 66.82 & 0.53 \\
  
  \midrule
  \multirow{3}{*}{ViT-L} & \IT clean performance & \IT 97.51 & \IT 76.94 & \IT 1.00 & \IT 0.42  & \IT 1.00 & \IT 88.91  & \IT 1.00 \\
  & low noise diffusion & 98.03 & 77.04 & 0.99 & 0.42 & 0.99 & 88.61 & 0.99 \\
  & high noise diffusion & 83.49 & 63.24 & 0.53 & 0.66  & 0.5  & 66.5 & 0.52 \\

  \bottomrule
  \end{tabular}
\end{table*}

\cref{tab:robustness_denoising} presents a comprehensive evaluation of DINOv2's performance across various backbone sizes, distortions, and adversarial scenarios. The results reveal several important patterns.

First, we observe a striking contrast between the model's resilience to common image distortions and its vulnerability to adversarial attacks. Focusing on the ViT-S variant, the foundation model exhibits remarkable robustness against practical transformations like horizontal flipping, Wiener filtering, blur, JPEG compression, and grayscale conversion. However, even a vanilla PGD adversarial attack devastates model performance across all metrics. This performance collapse underscores the critical vulnerability of foundation models to adversarial perturbations, even when these perturbations remain imperceptible to humans %

Regarding diffusion denoising, consistent patterns emerge across all model variants. Low-noise denoising preserves or slightly improves performance compared to unmodified inputs. %

However, high-noise diffusion significantly degrades performance across all tasks and model sizes. Interestingly, while larger models (ViT-B and ViT-L) demonstrate better baseline performance on clean images compared to ViT-S, the relative impact of diffusion denoising remains consistent across model scales. This suggests that the fundamental trade-off between noise intensity for potential adversarial robustness and maintaining task performance applies regardless of model capacity.

\Cref{fig:denoising_examples} shows how diffusion denoising affects classification and segmentation. Denoising introduces distortion that causes misclassifications. The segmentation maps clearly reveal the impact: for the boat image, the segmentation map becomes empty as the main object is removed, while for the cat image, the segmentation becomes excessively noisy, resulting in incorrect predictions.

\subsection{Diffusion During Attack and Defense}
\cref{tab:dinov2_adversarial_denoising} presents a comprehensive evaluation of the interaction between diffusion-based attacks and defenses. In this table, the first three columns specify the attack configuration (optimization method, diffusion timestep, and diffusion probability during attack), while the fourth column indicates the defense strategy. The remaining columns show performance metrics for PascalVOC, NYU Depth and rOxford datasets.

\paragraph{High Noise Diffusion Defense} as a defense strategy demonstrates remarkable effectiveness against all attack configurations, regardless of the attack method or diffusion settings during the attack. This consistent performance suggests that high noise diffusion successfully disrupts adversarial perturbations by substantially altering the input signal, effectively "resetting" the image features. It is important to note that this defense comes with a significant drawback: using high noise diffusion invariably results in a loss of approximately 20\% of the original model performance across all tasks, regardless of whether the input is clean or adversarial.

\paragraph{Low Noise Diffusion Defense} shows varying effectiveness depending on the attack configuration. When attackers incorporate diffusion during the attack process, particularly with low timesteps, the defense becomes significantly less effective. This suggests attackers can adapt to low noise defenses by incorporating similar noise levels during the attack process.

\paragraph{Diffusion During Attack.} The impact of incorporating diffusion during the attack process reveals interesting patterns. When attackers use high noise diffusion with $p_{\text{diffusion}}=1.0$, attacks become ineffective regardless of defense strategy, with performance metrics comparable to clean images. This counterintuitive result suggests that high noise diffusion during attack effectively destroys the adversarial perturbation itself. Conversely, using low noise diffusion or variable noise levels ([low, high]) during attack with $p_{\text{diffusion}}=1.0$ creates more transferable adversarial examples that maintain some effectiveness even against low noise defenses.

\paragraph{Attack Method Comparison.} Among the three attack methods, SIA consistently demonstrates greater resistance to diffusion-based defenses, particularly with low noise defense. For instance, with low noise defense and no diffusion during attack, SIA maintains lower classification accuracy (2.2\%) compared to PGD (46.3\%), indicating SIA generates more robust adversarial perturbations. This advantage diminishes with high noise defense, where all attack methods become comparably ineffective.

These results highlight a complex interplay between attack and defense strategies in the diffusion space. High noise diffusion provides strong defense but at the cost of significant performance degradation on clean inputs (as seen in \cref{tab:robustness_denoising}). Meanwhile, low noise diffusion preserves clean performance but remains vulnerable to sophisticated attacks that incorporate diffusion during the attack process. This creates a challenging trade-off between robustness and \perf that requires careful consideration based on application-specific requirements.

\begin{table*}[htbp]
\caption{Impact of diffusion denoising on DiNOv2 ViT-S model on PascalVOC validation set, NYU Depth test set and rOxford dataset \\ after different adversarial attacks. The arrow indicates the best performance for the defender. }
\label{tab:dinov2_adversarial_denoising} 
\centering

\begin{tabular}{ccc|c|ccc|cc|cc}
\toprule
\multicolumn{3}{c|}{\textbf{Attack}} & \textbf{Defense} & \multicolumn{3}{|c|}{\textbf{PascalVOC}} & \multicolumn{2}{c|}{\textbf{NYU Depth}} & \multicolumn{2}{c}{\textbf{rOxford}} \\

\textbf{Optim} & \textbf{Diff. t} & \textbf{Diff. prob} & \textbf{Denoising} & \textbf{Classif.$\uparrow$} & \textbf{Segment.$\uparrow$} & \textbf{\texttt{CLS} cos\_sim$\uparrow$} & \textbf{Depth$\downarrow$} & \textbf{\texttt{CLS} cos\_sim$\uparrow$} & \textbf{mAP$\uparrow$} & \textbf{\texttt{CLS} cos\_sim$\uparrow$} \\

\midrule
\multicolumn{3}{c|}{\multirow{3}{*}{None}} &None & 94.89 & 79.26 & 1.00 & 0.55 & 1.00 & 84.46 & 1.00 \\
&&&low noise & 95.54 & 79.03 & 0.99 & 0.58 & 0.99 & 83.82 & 0.99 \\
&&&high noise & 80.60 & 61.30 & 0.60 & 0.79 & 0.60 & 62.03 & 0.61 \\

\midrule

\multirow{7}{*}{pgd} & None & 0.0 & \multirow{21}{*}{None} & 0.0 & 10.5 & 0.2 & 6.4 & 0.1 & 0.46 & -0.62 \\
 & low & 0.5 & & 0.0 & 10.3 & 0.2 & 6.4 & 0.1 & 0.46 & -0.52 \\
 & low & 1.0 & & 0.0 & 11.7 & 0.2 & 6.2 & 0.2 & 0.47 & -0.28 \\
 & high & 0.5 & & 0.0 & 10.9 & 0.3 & 6.4 & 0.2 & 0.46 & -0.52 \\
 & high & 1.0 & & 93.3 & 76.9 & 1.0 & 0.6 & 1.0 & 84.84 & 0.99 \\
 & {[low, high]} & 0.5 & & 0.0 & 11.0 & 0.3 & 6.4 & 0.2 & 0.46 & -0.53 \\
 & {[low, high]} & 1.0 & & 30.3 & 45.2 & 0.85 & 2.1 & 0.8 & 73.99 & 0.78 \\
\cmidrule(lr){1-3} \cmidrule(lr){5-11}
\multirow{7}{*}{mifgsm} & None & 0.0 & & 0.0 & 13.5 & 0.25 & 6.3 & 0.2 & 0.47 & -0.39 \\
 & low & 0.5 & & 0.0 & 15.7 & 0.3 & 5.9 & 0.3 & 0.56 & -0.18 \\
 & low & 1.0 & & 0.0 & 20.4 & 0.4 & 5.1 & 0.4 & 8.81 & 0.08 \\
 & high & 0.5 & & 0.0 & 16.6 & 0.4 & 5.8 & 0.3 & 0.69 & -0.16 \\
 & high & 1.0 & & 92.9 & 77.0 & 1.0 & 0.6 & 1.0 & 85.13 & 0.99 \\
 & {[low, high]} & 0.5 & & 0.0 & 16.3 & 0.4 & 5.8 & 0.3 & 0.64 & -0.19 \\
 & {[low, high]} & 1.0 & & 37.6 & 49.9 & 0.85 & 1.8 & 0.9 & 80.47 & 0.83 \\
\cmidrule(lr){1-3} \cmidrule(lr){5-11}
\multirow{7}{*}{sia} & None & 0.0 & & 0.4 & 21.3 & 0.25 & 4.1 & 0.2 & 0.63 & -0.11 \\
 & low & 0.5 & & 1.2 & 23.6 & 0.35 & 4.1 & 0.3 & 1.55 & -0.02 \\
 & low & 1.0 & & 3.5 & 27.3 & 0.45 & 3.8 & 0.4 & 18.40 & 0.18 \\
 & high & 0.5 & & 3.3 & 27.5 & 0.4 & 3.8 & 0.3 & 3.95 & 0.02 \\
 & high & 1.0 & & 93.1 & 77.7 & 1.0 & 0.6 & 1.0 & 84.83 & 0.99 \\
 & {[low, high]} & 0.5 & & 2.6 & 26.0 & 0.35 & 3.8 & 0.3 & 4.50 & 0.03 \\
 & {[low, high]} & 1.0 & & 65.0 & 58.9 & 0.9 & 1.4 & 0.9 & 81.06 & 0.89 \\
\midrule
\multirow{7}{*}{pgd} & None & 0.0 & \multirow{21}{*}{low noise} & 46.3 & 54.6 & 0.9 & 1.4 & 0.9 & 81.66 & 0.88 \\
 & low & 0.5 & & 0.0 & 13.9 & 0.45 & 5.5 & 0.4 & 5.71 & 0.01 \\
 & low & 1.0 & & 0.0 & 12.4 & 0.35 & 6.1 & 0.3 & 0.53 & -0.19 \\
 & high & 0.5 & & 52.3 & 56.5 & 0.9 & 1.2 & 0.9 & 81.79 & 0.88 \\
 & high & 1.0 & & 93.1 & 76.6 & 1.0 & 0.6 & 1.0 & 84.35 & 0.98 \\
 & {[low, high]} & 0.5 & & 28.6 & 45.8 & 0.85 & 1.9 & 0.9 & 71.64 & 0.78 \\
 & {[low, high]} & 1.0 & & 34.2 & 45.9 & 0.85 & 2.1 & 0.8 & 74.51 & 0.77 \\
\cmidrule(lr){1-3} \cmidrule(lr){5-11}
\multirow{7}{*}{mifgsm} & None & 0.0 & & 9.2 & 34.6 & 0.8 & 2.5 & 0.8 & 50.07 & 0.55 \\
 & low & 0.5 & & 0.0 & 20.9 & 0.55 & 4.9 & 0.5 & 11.04 & 0.17 \\
 & low & 1.0 & & 0.1 & 21.9 & 0.5 & 4.9 & 0.5 & 18.99 & 0.17 \\
 & high & 0.5 & & 12.8 & 37.3 & 0.85 & 2.5 & 0.8 & 62.26 & 0.64 \\
 & high & 1.0 & & 93.6 & 76.5 & 1.0 & 0.7 & 1.0 & 84.92 & 0.98 \\
 & {[low, high]} & 0.5 & & 6.8 & 34.0 & 0.75 & 3.0 & 0.8 & 50.61 & 0.47 \\
 & {[low, high]} & 1.0 & & 40.2 & 50.0 & 0.85 & 1.8 & 0.9 & 80.01 & 0.82 \\
\cmidrule(lr){1-3} \cmidrule(lr){5-11}
\multirow{7}{*}{sia} & None & 0.0 & & 2.2 & 26.0 & 0.5 & 3.7 & 0.4 & 15.42 & 0.15 \\
 & low & 0.5 & & 3.7 & 27.0 & 0.45 & 3.7 & 0.4 & 18.43 & 0.17 \\
 & low & 1.0 & & 6.2 & 28.5 & 0.45 & 3.7 & 0.4 & 24.98 & 0.25 \\
 & high & 0.5 & & 12.7 & 34.1 & 0.65 & 3.1 & 0.5 & 36.79 & 0.35 \\
 & high & 1.0 & & 93.6 & 77.5 & 1.0 & 0.6 & 1.0 & 84.72 & 0.98 \\
 & {[low, high]} & 0.5 & & 9.6 & 32.1 & 0.6 & 3.2 & 0.5 & 35.14 & 0.32 \\
 & {[low, high]} & 1.0 & & 65.5 & 59.1 & 0.9 & 1.3 & 0.9 & 80.92 & 0.87 \\
\midrule
\multirow{7}{*}{pgd} & None & 0.0 & \multirow{21}{*}{high noise} & 79.4 & 61.5 & 0.6 & 0.8 & 0.6 & 64.55 & 0.60 \\
 & low & 0.5 & & 79.8 & 61.5 & 0.6 & 0.8 & 0.6 & 62.70 & 0.60 \\
 & low & 1.0 & & 79.8 & 61.0 & 0.6 & 0.8 & 0.6 & 59.80 & 0.61 \\
 & high & 0.5 & & 75.2 & 56.3 & 0.6 & 0.9 & 0.6 & 61.78 & 0.56 \\
 & high & 1.0 & & 64.5 & 52.4 & 0.6 & 1.1 & 0.5 & 54.58 & 0.52 \\
 & {[low, high]} & 0.5 & & 77.2 & 59.4 & 0.6 & 0.8 & 0.6 & 62.06 & 0.58 \\
 & {[low, high]} & 1.0 & & 73.1 & 57.7 & 0.6 & 0.9 & 0.6 & 56.71 & 0.56 \\
\cmidrule(lr){1-3} \cmidrule(lr){5-11}
\multirow{7}{*}{mifgsm} & None & 0.0 & & 79.0 & 61.3 & 0.6 & 0.8 & 0.6 & 66.18 & 0.59 \\
 & low & 0.5 & & 77.6 & 61.0 & 0.6 & 0.8 & 0.6 & 61.56 & 0.59 \\
 & low & 1.0 & & 79.8 & 61.1 & 0.6 & 0.8 & 0.6 & 66.63 & 0.62 \\
 & high & 0.5 & & 73.0 & 57.1 & 0.6 & 0.9 & 0.6 & 60.44 & 0.56 \\
 & high & 1.0 & & 67.0 & 53.6 & 0.6 & 1.0 & 0.5 & 60.42 & 0.52 \\
 & {[low, high]} & 0.5 & & 78.0 & 59.0 & 0.6 & 0.8 & 0.6 & 60.59 & 0.57 \\
 & {[low, high]} & 1.0 & & 72.9 & 57.7 & 0.6 & 0.9 & 0.6 & 55.52 & 0.57 \\
\cmidrule(lr){1-3} \cmidrule(lr){5-11}
\multirow{7}{*}{sia} & None & 0.0 & & 79.4 & 61.5 & 0.6 & 0.8 & 0.6 & 58.21 & 0.61 \\
 & low & 0.5 & & 78.6 & 60.4 & 0.6 & 0.8 & 0.6 & 63.34 & 0.58 \\
 & low & 1.0 & & 78.6 & 60.4 & 0.6 & 0.8 & 0.6 & 60.66 & 0.58 \\
 & high & 0.5 & & 72.0 & 56.9 & 0.6 & 0.9 & 0.6 & 57.40 & 0.55 \\
 & high & 1.0 & & 66.6 & 53.3 & 0.55 & 1.0 & 0.5 & 57.06 & 0.52 \\
 & {[low, high]} & 0.5 & & 77.3 & 58.7 & 0.6 & 0.8 & 0.6 & 57.77 & 0.58 \\
 & {[low, high]} & 1.0 & & 71.8 & 56.7 & 0.6 & 0.9 & 0.6 & 58.35 & 0.56 \\

\bottomrule
\end{tabular}
\end{table*}

\section{Conclusion}
This paper presents a comprehensive analysis of diffusion-based adversarial defenses for VFMs across multiple downstream tasks. Our experiments reveal key insights about robustness-performance trade-offs in DINOv2 models.

We observed a stark contrast between foundation models' resilience to common image distortions and their extreme vulnerability to adversarial attacks, with performance dropping from over 90\% to near-zero across multiple tasks. We identified that high-noise diffusion provides substantial protection against the tested attack methods but degrades clean performance by 14-33\% for classification, segmentation and retrieval tasks, and up to 57\% for depth estimation, while low-noise diffusion preserves performance but offers inadequate protection against sophisticated attacks.

Overall, while diffusion-based defenses offer promising protection for foundation models, the fundamental trade-off between robustness and \perf remains unresolved, presenting a significant challenge for deploying these models in security-critical contexts.

\bibliographystyle{IEEEtran} 
\bibliography{main}

\begin{thebibliography}{10}
\providecommand{\url}[1]{#1}
\csname url@samestyle\endcsname
\providecommand{\newblock}{\relax}
\providecommand{\bibinfo}[2]{#2}
\providecommand{\BIBentrySTDinterwordspacing}{\spaceskip=0pt\relax}
\providecommand{\BIBentryALTinterwordstretchfactor}{4}
\providecommand{\BIBentryALTinterwordspacing}{\spaceskip=\fontdimen2\font plus
\BIBentryALTinterwordstretchfactor\fontdimen3\font minus \fontdimen4\font\relax}
\providecommand{\BIBforeignlanguage}[2]{{%
\expandafter\ifx\csname l@#1\endcsname\relax
\typeout{** WARNING: IEEEtran.bst: No hyphenation pattern has been}%
\typeout{** loaded for the language `#1'. Using the pattern for}%
\typeout{** the default language instead.}%
\else
\language=\csname l@#1\endcsname
\fi
#2}}
\providecommand{\BIBdecl}{\relax}
\BIBdecl

\bibitem{carlini2023free}
N.~Carlini, F.~Tramèr, K.~Dvijotham, L.~Rice, M.~Sun, and Z.~Kolter, ``(certified!!) adversarial robustness for free!'' \emph{International Conference on Learning Representations (ICLR)}, 2023.

\bibitem{goodfellow2014explaining}
I.~J. Goodfellow, J.~Shlens, and C.~Szegedy, ``Explaining and harnessing adversarial examples,'' in \emph{3rd International Conference on Learning Representations, {ICLR} 2015, San Diego, CA, USA, May 7-9, 2015, Conference Track Proceedings}, Y.~Bengio and Y.~LeCun, Eds., 2015.

\bibitem{szegedy2013intriguing}
C.~Szegedy, W.~Zaremba, I.~Sutskever, J.~Bruna, D.~Erhan, I.~J. Goodfellow, and R.~Fergus, ``Intriguing properties of neural networks,'' in \emph{2nd International Conference on Learning Representations, {ICLR} 2014, Banff, AB, Canada, April 14-16, 2014, Conference Track Proceedings}, Y.~Bengio and Y.~LeCun, Eds., 2014.

\bibitem{madry18pgd}
A.~Madry, A.~Makelov, L.~Schmidt, D.~Tsipras, and A.~Vladu, ``Towards deep learning models resistant to adversarial attacks,'' in \emph{6th International Conference on Learning Representations, {ICLR} 2018, Vancouver, BC, Canada, April 30 - May 3, 2018, Conference Track Proceedings}.\hskip 1em plus 0.5em minus 0.4em\relax OpenReview.net, 2018.

\bibitem{dong2018mifgsm}
Y.~Dong, F.~Liao, T.~Pang, H.~Su, J.~Zhu, X.~Hu, and J.~Li, ``Boosting adversarial attacks with momentum,'' in \emph{Proceedings of the IEEE conference on computer vision and pattern recognition}, 2018, pp. 9185--9193.

\bibitem{wang2023sia}
X.~Wang, Z.~Zhang, and J.~Zhang, ``Structure invariant transformation for better adversarial transferability,'' in \emph{Proceedings of the IEEE/CVF International Conference on Computer Vision}, 2023, pp. 4607--4619.

\bibitem{guo2017countering}
C.~Guo, M.~Rana, M.~Cisse, and L.~Van Der~Maaten, ``Countering adversarial images using input transformations,'' \emph{arXiv preprint arXiv:1711.00117}, 2017.

\bibitem{cohen2019certified}
J.~Cohen, E.~Rosenfeld, and Z.~Kolter, ``Certified adversarial robustness via randomized smoothing,'' in \emph{international conference on machine learning}.\hskip 1em plus 0.5em minus 0.4em\relax PMLR, 2019, pp. 1310--1320.

\bibitem{salman2020denoised}
H.~Salman, M.~Sun, G.~Yang, A.~Kapoor, and J.~Z. Kolter, ``Denoised smoothing: A provable defense for pretrained classifiers,'' \emph{Advances in Neural Information Processing Systems}, vol.~33, pp. 21\,945--21\,957, 2020.

\bibitem{ho2020denoising}
J.~Ho, A.~Jain, and P.~Abbeel, ``Denoising diffusion probabilistic models,'' \emph{Advances in neural information processing systems}, vol.~33, pp. 6840--6851, 2020.

\bibitem{oquab2023dinov2}
M.~Oquab, T.~Darcet, T.~Moutakanni, H.~Vo, M.~Szafraniec, V.~Khalidov, P.~Fernandez, D.~Haziza, F.~Massa, A.~El-Nouby, M.~Assran, N.~Ballas, W.~Galuba, R.~Howes, P.-Y. Huang, S.-W. Li, I.~Misra, M.~Rabbat, V.~Sharma, G.~Synnaeve, H.~Xu, H.~Jegou, J.~Mairal, P.~Labatut, A.~Joulin, and P.~Bojanowski, ``Dinov2: Learning robust visual features without supervision,'' 2023.

\bibitem{dosovitskiy2021image}
A.~Dosovitskiy, L.~Beyer, A.~Kolesnikov, D.~Weissenborn, X.~Zhai, T.~Unterthiner, M.~Dehghani, M.~Minderer, G.~Heigold, S.~Gelly, J.~Uszkoreit, and N.~Houlsby, ``An image is worth 16x16 words: Transformers for image recognition at scale,'' 2021.

\bibitem{dhariwal2021diffusion}
P.~Dhariwal and A.~Nichol, ``Diffusion models beat gans on image synthesis,'' \emph{Advances in neural information processing systems}, vol.~34, pp. 8780--8794, 2021.

\bibitem{deng2009imagenet}
J.~Deng, W.~Dong, R.~Socher, L.-J. Li, K.~Li, and L.~Fei-Fei, ``Imagenet: A large-scale hierarchical image database,'' in \emph{2009 IEEE Conference on Computer Vision and Pattern Recognition}, 2009, pp. 248--255.

\bibitem{pascal-voc-2012}
M.~Everingham, L.~Van~Gool, C.~K.~I. Williams, J.~Winn, and A.~Zisserman, ``The {PASCAL} {V}isual {O}bject {C}lasses {C}hallenge 2012 {(VOC2012)} {R}esults,'' http://www.pascal-network.org/challenges/VOC/voc2012/workshop/index.html.

\bibitem{couprie2013indoor}
C.~Couprie, C.~Farabet, L.~Najman, and Y.~LeCun, ``Indoor semantic segmentation using depth information,'' \emph{arXiv preprint arXiv:1301.3572}, 2013.

\bibitem{RITAC18}
F.~Radenovi\'{c}, A.~Iscen, G.~Tolias, Y.~Avrithis, and O.~Chum, ``Revisiting oxford and paris: Large-scale image retrieval benchmarking,'' in \emph{CVPR}, 2018.

\end{thebibliography}

\end{document}